\documentclass[letterpaper, 10 pt, conference]{ieeeconf}  
\IEEEoverridecommandlockouts
\usepackage{cite}
\usepackage{amsmath,amssymb,amsfonts}
\usepackage{algorithm}
\usepackage{booktabs} 
\usepackage{algpseudocode}
\usepackage{graphicx}
\usepackage{textcomp}
\usepackage{xcolor}
\usepackage{subfiles}
\usepackage{mathtools}
\usepackage{subcaption}
\usepackage{graphicx}
\usepackage{svg}
\usepackage{url}
\usepackage{hyperref}

\def\BibTeX{{\rm B\kern-.05em{\sc i\kern-.025em b}\kern-.08em
    T\kern-.1667em\lower.7ex\hbox{E}\kern-.125emX}}

\begin{document}
\title{\LARGE \bf MIXED-SENSE: A Mixed Reality Sensor Emulation Framework for Test and Evaluation of UAVs Against False Data Injection Attacks}
\author{Kartik A. Pant, Li-Yu Lin, Jaehyeok Kim, Worawis Sribunma, James M. Goppert, Inseok Hwang
\thanks{This research is funded by the Secure Systems Research Center (SSRC) at the Technology Innovation Institute (TII), UAE. The authors are grateful to Dr. Shreekant (Ticky) Thakkar and his team members at the SSRC for their valuable comments and support.}
\thanks{The authors are with the School of Aeronautics and Astronautics, Purdue University,
West Lafayette, IN 47906. Email: ({\tt\small kpant@purdue.edu}, {\tt\small lin1191@purdue.edu}, {\tt\small kim2153@purdue.edu}, {\tt\small wsribunm@purdue.edu}, {\tt\small jgoppert@purdue.edu}, {\tt\small ihwang@purdue.edu})}%
}

\maketitle
\begin{abstract}
We present a high-fidelity Mixed Reality sensor emulation framework for testing and evaluating the resilience of Unmanned Aerial Vehicles (UAVs) against false data injection (FDI) attacks. 
The proposed approach can be utilized to assess the impact of FDI attacks, benchmark attack detector performance, and validate the effectiveness of mitigation/reconfiguration strategies in single-UAV and UAV swarm operations. 
Our Mixed Reality framework leverages high-fidelity simulations of Gazebo and a Motion Capture system to emulate proprioceptive (e.g., GNSS) and exteroceptive (e.g., camera) sensor measurements in real-time. We propose an empirical approach to faithfully recreate signal characteristics such as latency and noise in these measurements. 
%
Finally, we illustrate the efficacy of our proposed framework through a Mixed Reality experiment consisting of an emulated GNSS attack on an actual UAV, which (i) demonstrates the impact of false data injection attacks on GNSS measurements and (ii) validates a mitigation strategy utilizing a distributed camera network developed in our previous work. Our open-source implementation is available at \href{https://github.com/CogniPilot/mixed\_sense}{\texttt{https://github.com/CogniPilot/mixed\_sense}}
\end{abstract}

%
%
%
%
%
%

\section{Introduction}
\label{sec:intro}
Recent years have seen a proliferation of UAVs (autonomous and semi-autonomous) for various applications such as search and rescue, disaster management,  agriculture, etc. While these aerial vehicles perform their mission, they are often exposed to security threats imposed by adversaries to disrupt their operations. For instance, the recent events in the Russia-Ukraine war have demonstrated the use of cyber-warfare systems to neutralize drones actively by jamming and spoofing GNSS measurements \cite{ukrainerussia2024}. These threats often result from vulnerabilities exhibited by sensors and communication channels of UAVs \cite{krishna2017review}. An intelligent adversary can manipulate the UAV's behavior to its advantage by injecting false data in either the sensor feed, e.g.,  GNSS spoofing \cite{kerns2014unmanned, psiaki2016gnss}, or can disrupt the communication links, e.g., Command-and-Control (C2) and Vehicle-to-Vehicle (V2V) link hijacking \cite{kwon2018empirical}, to take control over the system.
\begin{figure}[ht]
\centering
\includegraphics[width=0.42\textwidth]{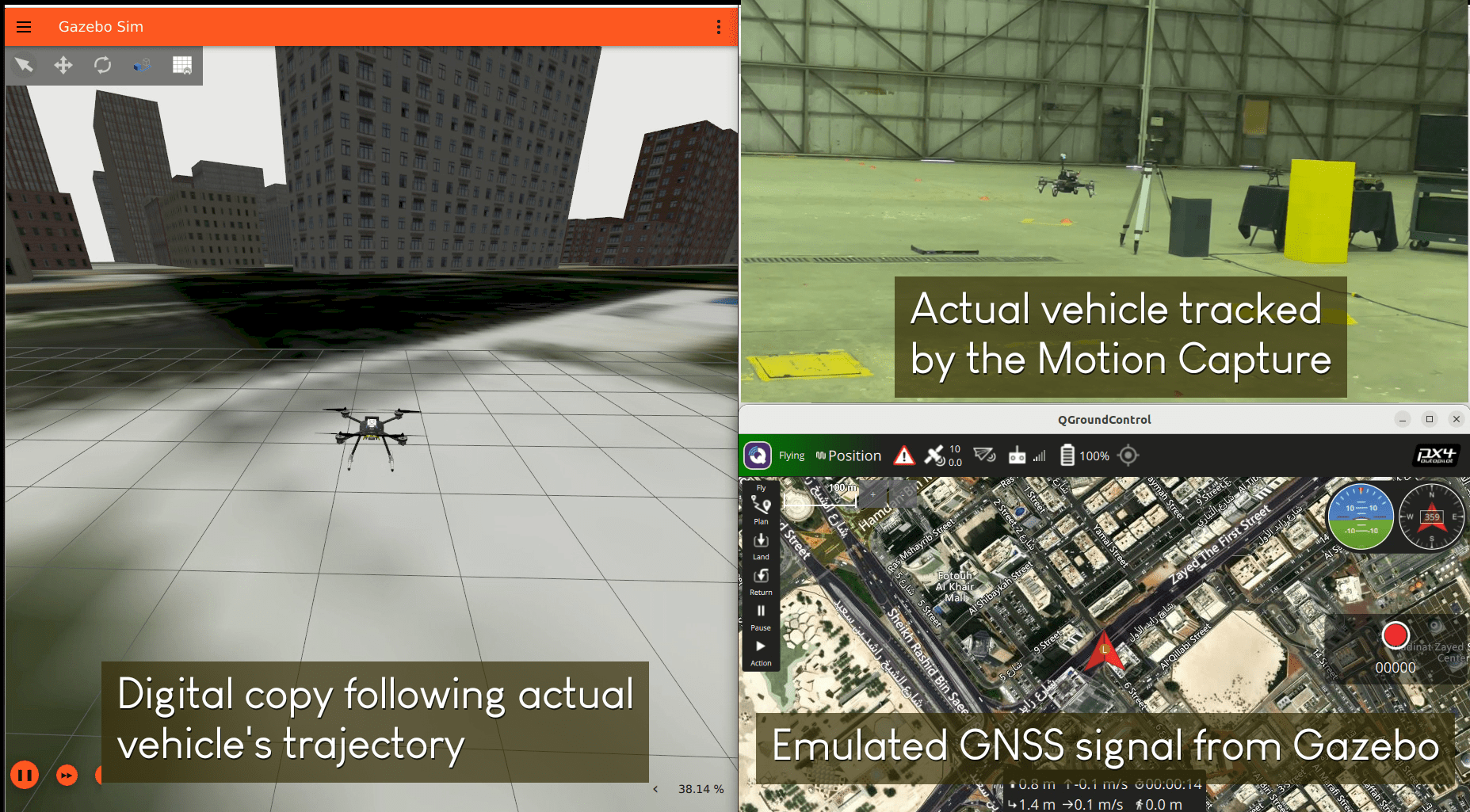}
\caption{An instance of Mixed-Reality-in-the-loop (MRiTL) GNSS sensor emulation}
\label{fig:mixed_reality}
\end{figure}

Test and evaluation of UAVs before deployment is essential to assess the impact of such adversarial false data injection attacks. This allows us to quickly identify and eliminate the effects of the system's vulnerabilities. The current methods for test and evaluation rely either on software-in-the-loop (SiTL) tests (e.g., PX4 SiTL \cite{px4ekf2020}) or controlled hardware tests (e.g., controlled over-the-air GNSS spoofing tests \cite{sathaye2022experimental}). One of the major drawbacks of such approaches is their lower level of fidelity, which limits their use for cybersecurity applications. To this end, we present a Mixed Reality (MR) sensor emulation framework to analyze and mitigate the impact of false data injection attacks on UAVs. Mixed Reality has emerged as a new paradigm for vehicle-in-the-loop simulation, an extension of the traditional hardware-in-the-loop (HiTL) simulation. It represents the spectrum encompassing both Augmented Reality (AR) and Virtual Reality (VR), synergistically combining elements of the physical and digital world.  

The rise of MR technology has been fueled by recent advancements in the hardware, e.g., advanced graphic processors with GPU acceleration, and software frameworks, e.g., gaming platforms such as Unreal Engine and Unity, that render high-fidelity photorealistic simulations in real time. Several datasets, Synthia \cite{ros2016synthia}, Virtual KITTI \cite{gaidon2016virtual}, and Blackbird \cite{antonini2020blackbird}, comprising photorealistic renderings of urban environments, are being actively used in robotics research. Another major driving force in the maturation of Mixed Reality simulations is the advancements in Motion Capture (Mocap) technologies, i.e., infrared cameras, laser tracking, etc. Motion Capture facilitates the incorporation of natural motion and behavior of actual vehicles into the simulation, along with its interaction with virtual components in real-time. 
\begin{figure*}[ht]
  \centering
  \includegraphics[width=0.8\textwidth]{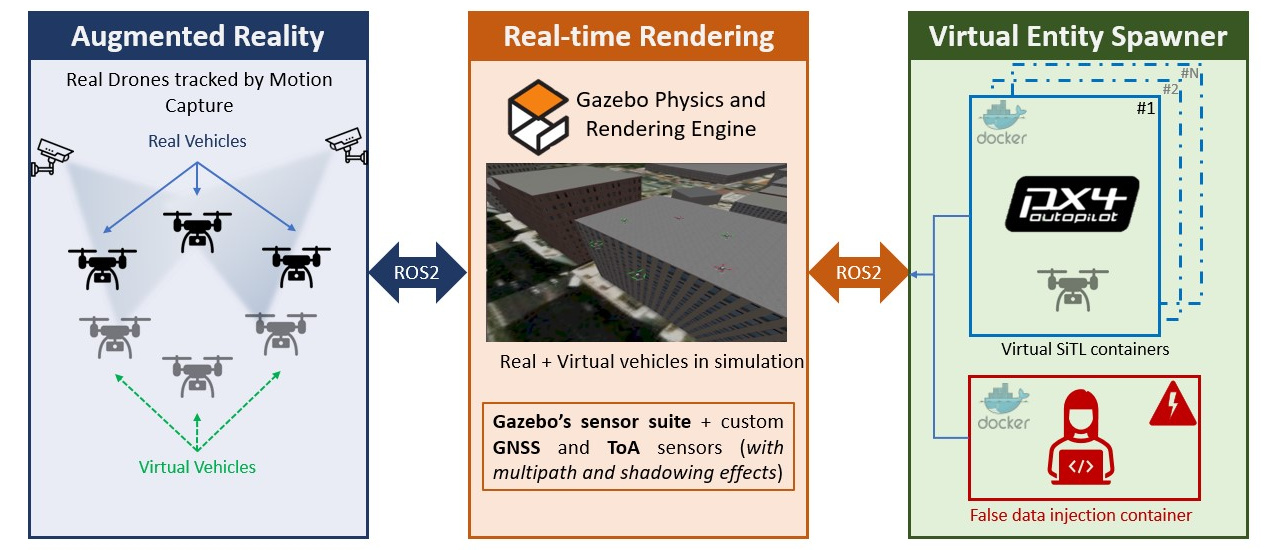}
  \caption{Overview of our proposed framework. Gazebo's rendering engine spawns real vehicles tracked in the motion capture and virtual vehicles running as PX4 SiTL instances. Gazebo's physics engine performs the real-time processing of scene rendering, sensor emulation, and collision detection. 
  }
  \label{fig:arch}

\end{figure*}

MR can play a vital role in UAV cybersecurity analysis. For example, conducting a GNSS spoofing experiment in a densely populated urban environment, e.g., Downtown Chicago, can be practically infeasible as it would disrupt the existing operations pertaining to commercial aviation, military, autonomous cars, and air taxis, which rely heavily on GNSS measurements.
However, with improved sensor modeling and photorealistic urban environment rendering, one can utilize Mixed-Reality-in-the-loop (MRiTL) experiments to characterize the impact of such attacks without actually flying the vehicle in the urban canyon. Furthermore, a Mixed Reality sensor emulation framework can be leveraged to conduct more sophisticated false data injection (FDI) attacks that simultaneously impact multiple sensors and communication channels. Thus, our proposed framework provides the potential to combine the effectiveness, safety, and adaptability of simulation with real-world physics for UAV cybersecurity applications. 

In this paper, our main contributions are as follows.
\begin{enumerate}
    \item We propose a Mixed Reality sensor emulation framework to test and evaluate UAVs against FDI attacks.
    \item We analyze the effects of latency in sensor emulation on MRiTL experiments and provide a systematic approach to tune the emulated sensors' latency empirically.
    \item We experimentally demonstrate the effectiveness of our proposed framework on an actual UAV by assessing the impact of FDI attacks on GNSS measurements and validating a mitigation strategy utilizing external position measurements from a distributed camera network to detect and mitigate the impact of such attacks.
\end{enumerate}

The rest of the paper is organized as follows. Section \ref{sec:related_work} summarizes related works in MR for robotics and UAVs and the existing literature on testbed development for UAV cybersecurity. In Section \ref{sec:sys_arch}, we describe the system architecture of the proposed framework, including the interfacing of emulated sensors from the simulator with the actual test vehicles. Section \ref{sec:latency} outlines the effects of latency in sensor emulation. This section also presents an empirical approach for tuning the emulated sensor latency for Mixed Reality experiments. Section \ref{sec:exp} describes the experimental validation of the proposed framework on an actual test vehicle by conducting an emulated GNSS replay attack (also known as GNSS meaconing attack) and validating the mitigation approach using external measurements. Finally, Section \ref{sec:conclusion} concludes the paper and presents future directions.

\section{Related Works}
\label{sec:related_work}
The idea of Mixed Reality goes back to 1994 \cite{milgram1994taxonomy}, where it was first described as a continuum between reality and virtuality, i.e., the merging of physical and virtual objects and environments. Since then, this idea has been widely adopted in various fields of research such as manufacturing\cite{malik2020nteract}, robotics \cite{hoenig2015mixed}, and automotive \cite{drechsler2022mire}. This section will mainly focus on the recent advances in Mixed Reality for robotics and UAV applications. We also present a brief survey of various testbed developments for UAV cybersecurity. 

\subsection{Mixed Reality in Robotics}
The recent advancements in 3D graphic rendering have enabled advanced graphic features such as material shading, real-time reflections, volumetric lighting, etc. These developments allow modern robotics simulators to achieve photorealistic rendering of scenes within simulation settings. Gazebo \cite{koenig2004design}, known for its tight integration with the Robot Operating System (ROS), has recently been upgraded to support high-fidelity scene rendering and physics. The Unreal engine, developed initially for game development, has gained popularity in robotics because of its high-fidelity scene rendering capabilities. Based on these advancements in high-fidelity simulation frameworks, several research efforts have been made to incorporate AR and VR technologies for robotics applications \cite{makhataeva2020augmented}. There have been efforts to combine AR/VR to develop Mixed Reality frameworks for robotics \cite{hoenig2015mixed, chen2009mixed}. The closest work to our approach is FlightGoggles \cite{guerra2019flightgoggles}, which utilizes Unreal engine and photogrammetry techniques to generate high-fidelity photorealistic simulations for UAVs. However, their work is limited to sensor emulation of exteroceptive sensors, e.g., cameras, IR beacons, etc. They do not provide sensor emulation of proprioceptive sensors, e.g., GNSS, etc, which are more susceptible to FDI attacks. 
\begin{figure}[t]
  \centering
  \includegraphics[width=0.47\textwidth]{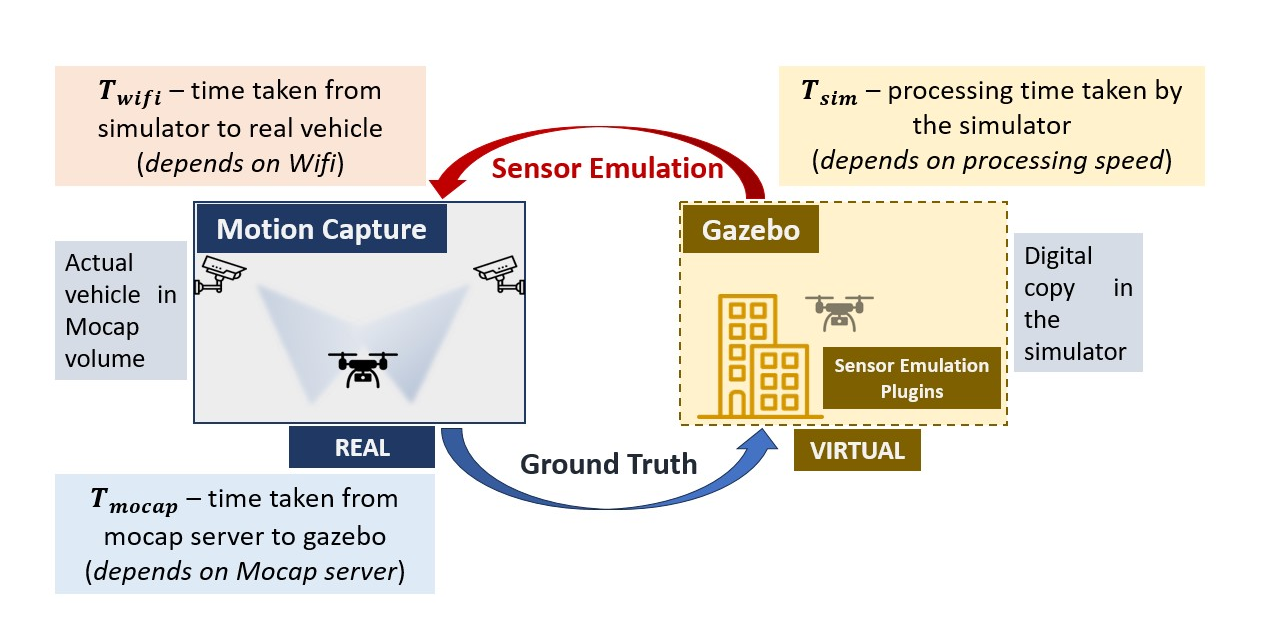}
  \caption{End-to-end time delay from sensing the motion of the real vehicle to publishing the emulated sensor measurements.}
  \label{fig:latency}
\end{figure}
\subsection{Testbed for UAV Cybersecurity}
Several testbeds have been developed for UAV applications. 
In \cite{schmittle2018openuav}, the authors have developed an open-source cloud-based testbed specifically for UAVs. It is based on Gazebo and PX4 Autopilot; however, the main focus was establishing a cloud-based platform for UAV research. The authors in \cite{javaid2013uavsim} proposed a simulation platform focusing on the cybersecurity analysis of UAVs. A SiTL and HiTL testbed for UAV cybersecurity was proposed in \cite{goppert2014software}; however, it lacks the fidelity to account for realistic scene rendering and sensor emulation. The authors in \cite{siewert2019fail} presented an experimental approach for fail-secure and fail-safe testing of UAVs and Urban Air Mobility (UAM) systems. However, it is targeted explicitly for false data injection attacks on GNSS measurements and Automatic Dependent Surveillance–Broadcast (ADS-B). Most existing approaches either rely purely on simulations (lacking the required level-of-fidelity) or hardware tests targeted for specific system vulnerabilities, making them ineffective for an end-to-end test and evaluation of UAVs against FDI attacks. To this end, our proposed framework enables a comprehensive cybersecurity analysis of UAVs, combining the real-world physics of UAVs with high-fidelity sensor emulation.


\section{Mixed Reality Sensor Emulation}
\label{sec:sys_arch}
Our sensor emulation framework follows a modular architecture, as shown in Fig. \ref{fig:arch}. The three major components of the framework are (i) Augmented Reality, (ii) Real-time Rendering, and (iii) Virtual Entity Spawner. Our architecture allows the user to modify the simulation scenario by combining real and virtual components as per the test specification. 
The core component of our framework is the Gazebo's physics engine, i.e., Bullet, and its rendering engine, i.e., Ogre2, which performs real-time processing of scene rendering, sensor emulation, and collision checking. Contrary to previous approaches \cite{guerra2019flightgoggles}, which only provide exteroceptive sensor models (such as cameras, IR  beacons, and time-of-flight range sensors), in this work, we extend the sensor emulation capabilities by also including GNSS \cite{pant2023open} and Time of Arrival (ToA) sensors (incorporating multipath and line-of-sight effects). We use Gazebo's improved rendering capabilities to generate photorealistic renderings of the simulation environment by importing high-resolution meshes from Google Earth 3D tiles \cite{googleearth}. Similar to \cite{guerra2019flightgoggles}, our framework allows users to import the meshes generated using photogrammetry into the simulation environment. All the information related to real and virtual vehicles is combined by Gazebo and made available as ROS topics. 
The Augmented Reality module consists of Motion Capture cameras tracking the motion of actual vehicles. The position and orientation data obtained via the Motion Capture system is transported to Gazebo, where digital avatars of the actual vehicles are created. These avatars are dragged inside the simulation environment following the same trajectory as the actual vehicle. The emulated sensor measurements in Gazebo are broadcasted to the actual test vehicles over the ROS2 network. 
The Virtual Entity Spawner module generates virtual vehicles in the simulation. Each virtual vehicle is created as a docker container instance of PX4 Autopilot firmware, replicating real-world communication and memory constraints. This allows each virtual vehicle to have independent resource allocation, memory management, and networking. The virtual vehicle uses Gazebo's physics engine for its simulated dynamics, collision checking, and inertial measurements. 

The main advantage of this approach stems from actual test vehicles experiencing environmental effects similar to those of an actual flight, e.g., multipath effects near high-rise buildings, etc., all within a controlled environment. It also allows the actual vehicle to interact with virtual components in a synchronized manner. As a result, the proposed method enables the discovery of vulnerabilities that can result from (i) environmental changes, e.g., weather, lighting, etc., adjusted via high fidelity simulations, (ii) unmodelled dynamics, e.g., aerodynamics, battery electrochemistry, motor vibrations, integrated through vehicle's natural dynamics, and (iii) interaction of actual and virtual vehicles, e.g., communication delays in C2 and V2V link, etc.  

\section{Latency in Sensor Emulation}
\label{sec:latency}
One of the significant challenges in MRiTL experiments is the latency in sensor emulation \cite{schnierle2023latency}. The latency is attributed to the rendering, data processing, and network overhead caused by the distributed components involved in the sensor emulation pipeline, i.e., the Motion Capture system, the ROS2 network, and the wireless connection to the actual vehicle. 
It is crucial to faithfully recreate signal characteristics such as latency and noise in the emulated sensor measurements, as the effectiveness of the FDI attack often relies upon such signal parameters. To achieve an accurate representation, the end-to-end latency in sensor emulation must follow a latency distribution similar to that of an actual sensor. 
In this work, we aim to identify the nominal latency caused by each component involved in the sensor emulation. This provides an empirical approach to tuning the end-to-end time delay\footnote{A more detailed theoretical analysis of the latency distribution of each sensor and compensation techniques to minimize its impact on sensor emulation will be considered in our future work.}. 
Figure \ref{fig:latency} represents the critical components responsible for latency in sensor emulation using our proposed framework. The Motion Capture system tracks the actual vehicle and processes the information to generate its 6 DoF pose. Let $T_{\text{mocap}}$ be the time a Motion Capture system takes to process the raw images to obtain a 6 DoF pose and transmit it over the network. Let $T_{\text{sim}}$ denote the time the Gazebo simulator takes to render the sensor in simulation and process its data. Finally, let $T_{\text{wifi}}$ denote the combined time taken by the Gazebo transport and the ROS2 network to actual vehicles flying in the Motion Capture volume. Thus, the total end-to-end time delay $T_{\text{total}}$ in sensor emulation for a particular sensor can be represented as:
\begin{equation}
    T_{\text{total}} = T_{\text{mocap}} + T_{\text{sim}} + T_{\text{wifi}}
\end{equation}
Most commercially available autopilots handle sensor delays as part of their state estimation logic. For example, PX4 Autopilot uses an extended Kalman filter (EKF) for state estimation. The compensation for sensor delay is incorporated in the EKF's update step. However, the nominal sensor delay must be specified as part of the UAV's configuration settings. This allows the EKF to compensate for the effects of time delay while fusing measurements from various onboard sensors. 
To this end, we propose to add a constant delay $T_{\text{c}}$ on top of the end-to-end time delay $ T_{\text{total}}$ to match the latency distribution of the actual sensor\footnote{Note that such an approach is only feasible if the end-to-end latency for an emulated sensor is less than the latency of the actual sensor.}. We find this empirical tuning approach effective through our experimental evaluation.   

\section{Experimental Demonstration}
\label{sec:exp}
\begin{figure}[ht]
\centering
\includegraphics[width = 0.43\textwidth]{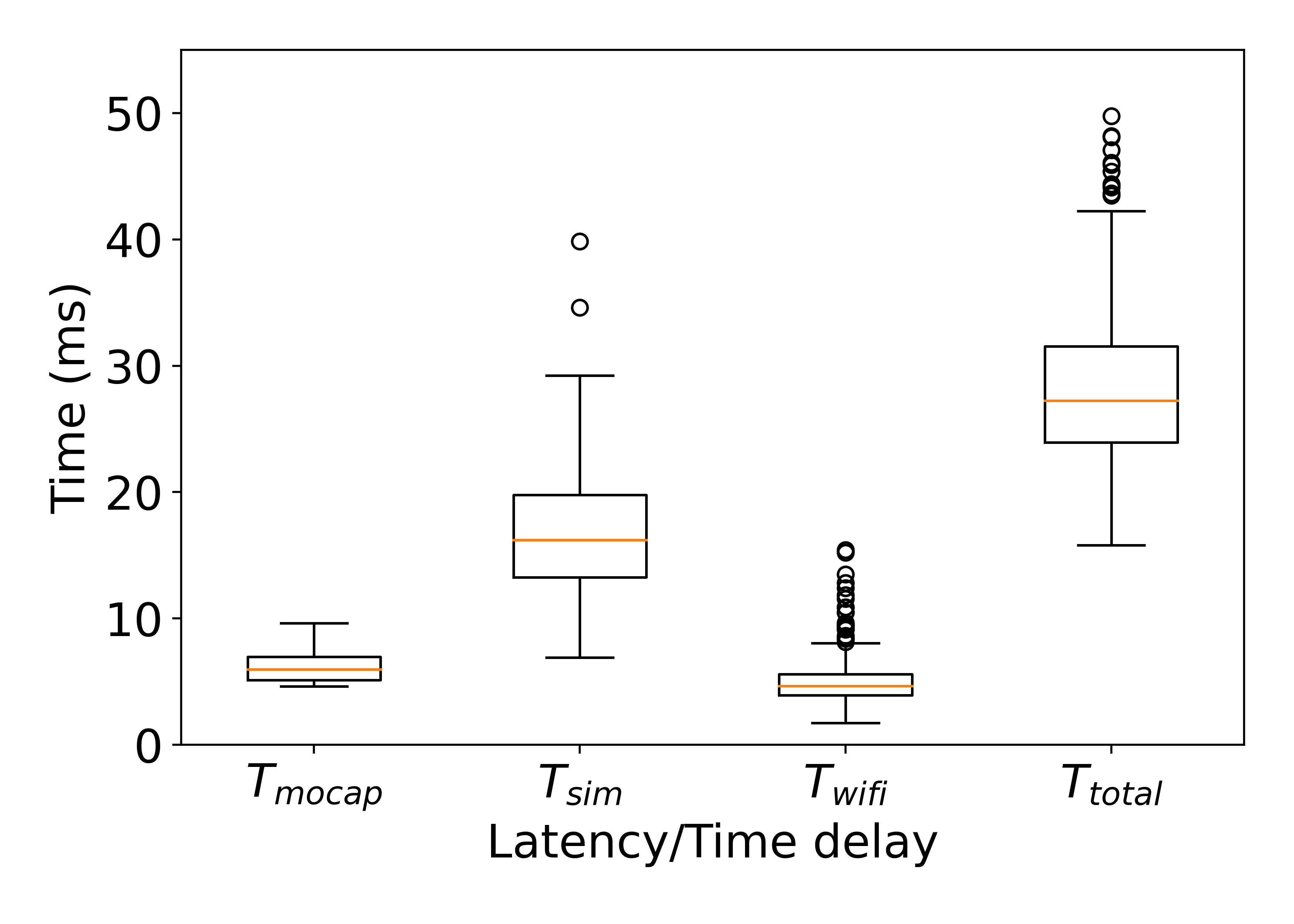}
\caption{Empirical latency in each component and total end-to-end latency in GNSS sensor emulation}
\label{fig:latency_box}
\end{figure}

This section describes the experimental demonstration of the proposed framework on an actual test vehicle in the MRiTL experiment. We emulate a GNSS meaconing attack to showcase the effectiveness of the proposed framework. We use a PX4 Vision drone with the PX4 Autopilot firmware as our test vehicle for experiments. The drone has a Pixhawk 6C flight controller, an UP Core companion computer, and a depth camera. We conduct our experiments at the Purdue UAS Research and Test Facility (PURT) which is the world's largest indoor motion capture facility. It comprises 60 Oqus 7+ Qualisys motion capture cameras distributed across 20,000 square feet and with a 30-foot ceiling. The PX4 Vision drone is interfaced with emulated GNSS sensor measurements from Gazebo. 
We use a custom GNSS sensor plugin \cite{pant2023open} which utilizes ray tracing to approximate multipath effects due to reflections from buildings and other structures rendered in the simulation. The GNSS sensor measurements are emulated using a Dell Precision-3581 workstation laptop with a 13{th}-Gen Intel Core i7 processor and an Nvidia RTX A1000 GPU. Figure \ref{fig:latency_box} depicts the empirical value of latency in various components recorded from a 2-minute GNSS sensor emulation experiment. The nominal latency value for the Qualysis Motion Capture system with 60 Oqus 7+ cameras is 6.02 $\pm$ 0.88 ms, the processing time taken by Gazebo has a nominal value of 16.01 $\pm$ 4.45 ms, and the ping latency (one-side) of the wifi network has a nominal value of 4.62 $\pm$ 0.98 ms. We add a delay of 73 ms to the nominal end-to-end latency of 27.42 $\pm$ 4.11 ms to match the latency value of a typical uBlox receiver, i.e., 100 ms, set in the PX4Autopilot's configuration file \cite{px4ekf2020}.   

\begin{figure*}[ht]
    \centering
    \begin{subfigure}[t]{0.47\textwidth}
      \centering
      \includegraphics[width=\textwidth,  height = 0.52\textwidth]{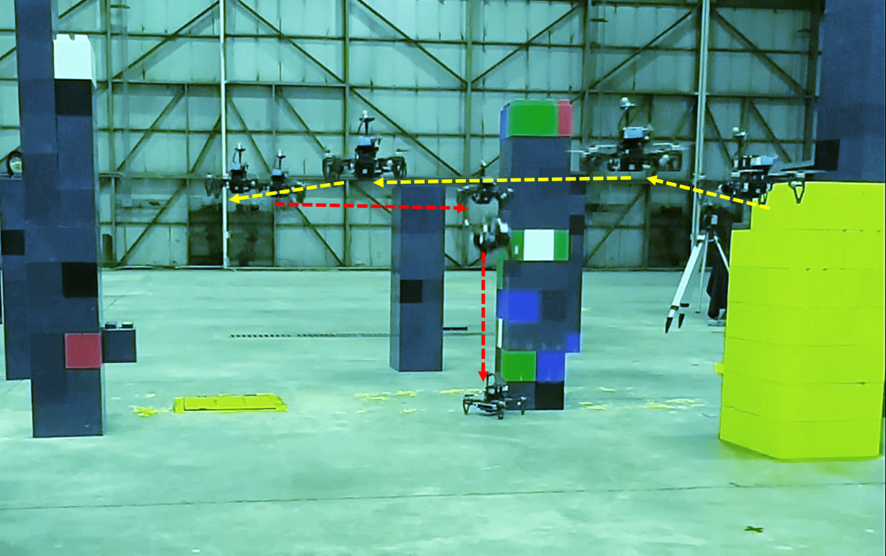}
      \caption{Emulated GNSS meaconing attack causing the actual vehicle to crash without triggering a failsafe.}
      \label{fig:experiment_a}
    \end{subfigure}
    \hfill
    \begin{subfigure}[t]{0.47\textwidth}
      \centering
      \includegraphics[width=\textwidth,  height = 0.52\textwidth]{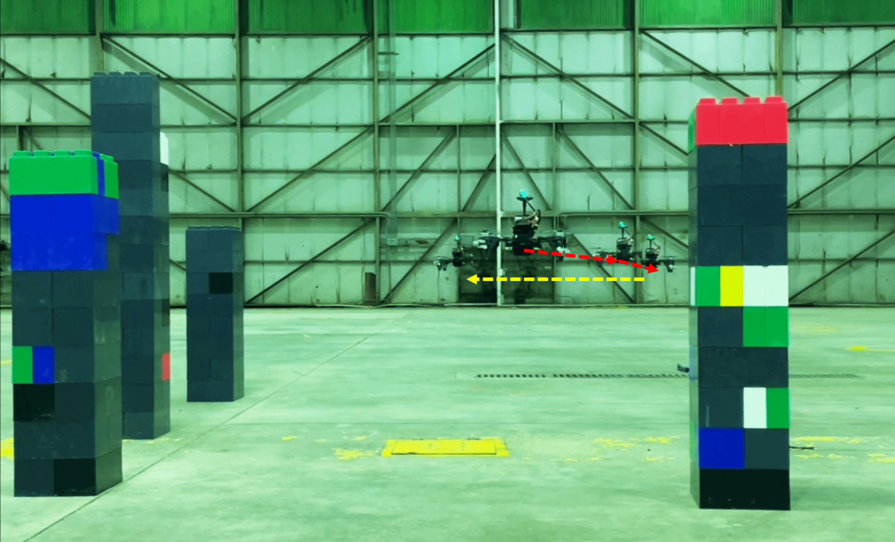}
      \caption{Vehicle fuses external position along with GNSS measurements to correct its position once the attack is detected.}
      \label{fig:experiment_b}
    \end{subfigure}
    
    \begin{subfigure}[t]{\textwidth}
      \centering
      \includegraphics[width=0.47\textwidth, height = 0.42\textwidth]{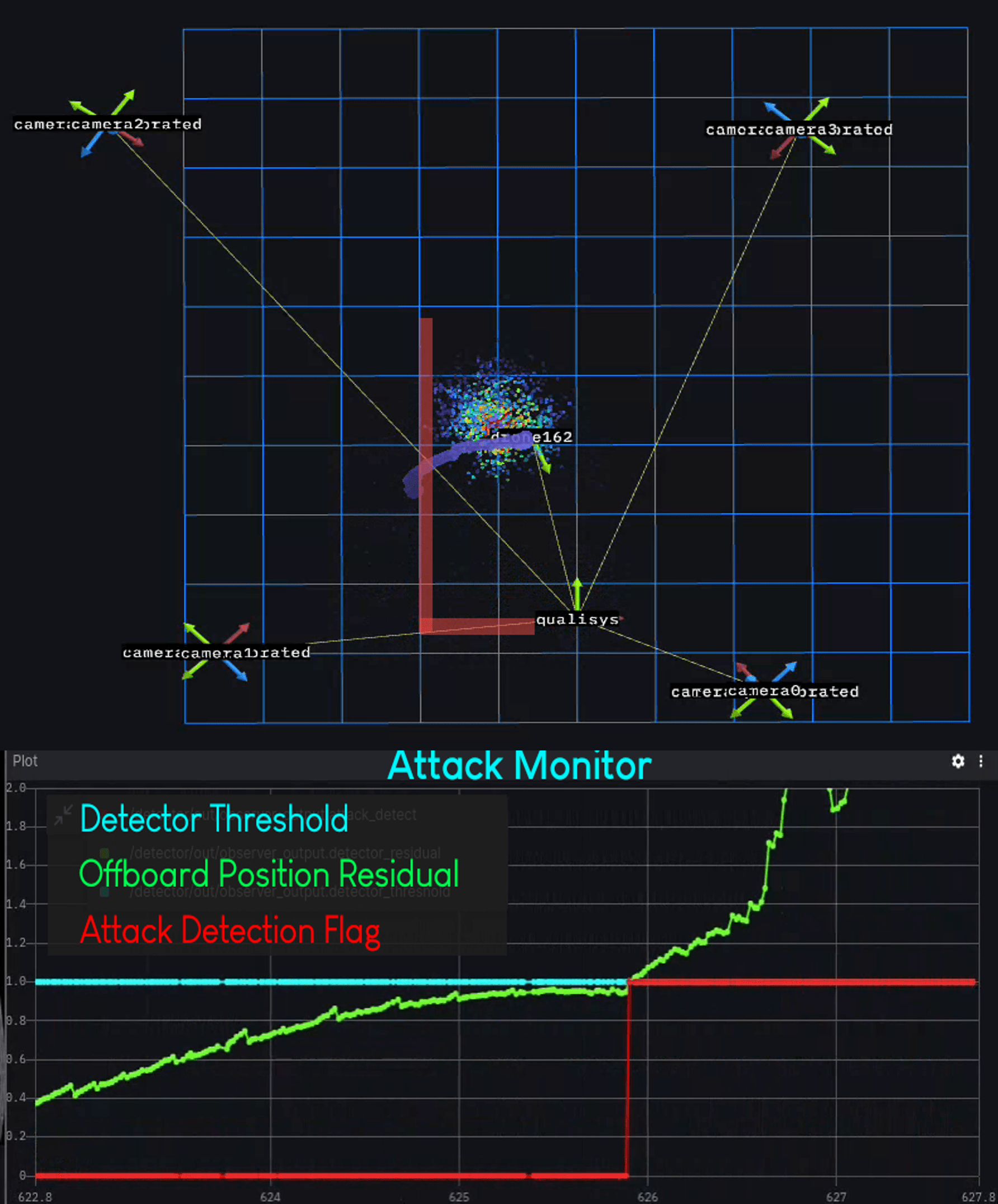}
      \hfill
      \includegraphics[width=0.47\textwidth, height = 0.42\textwidth]{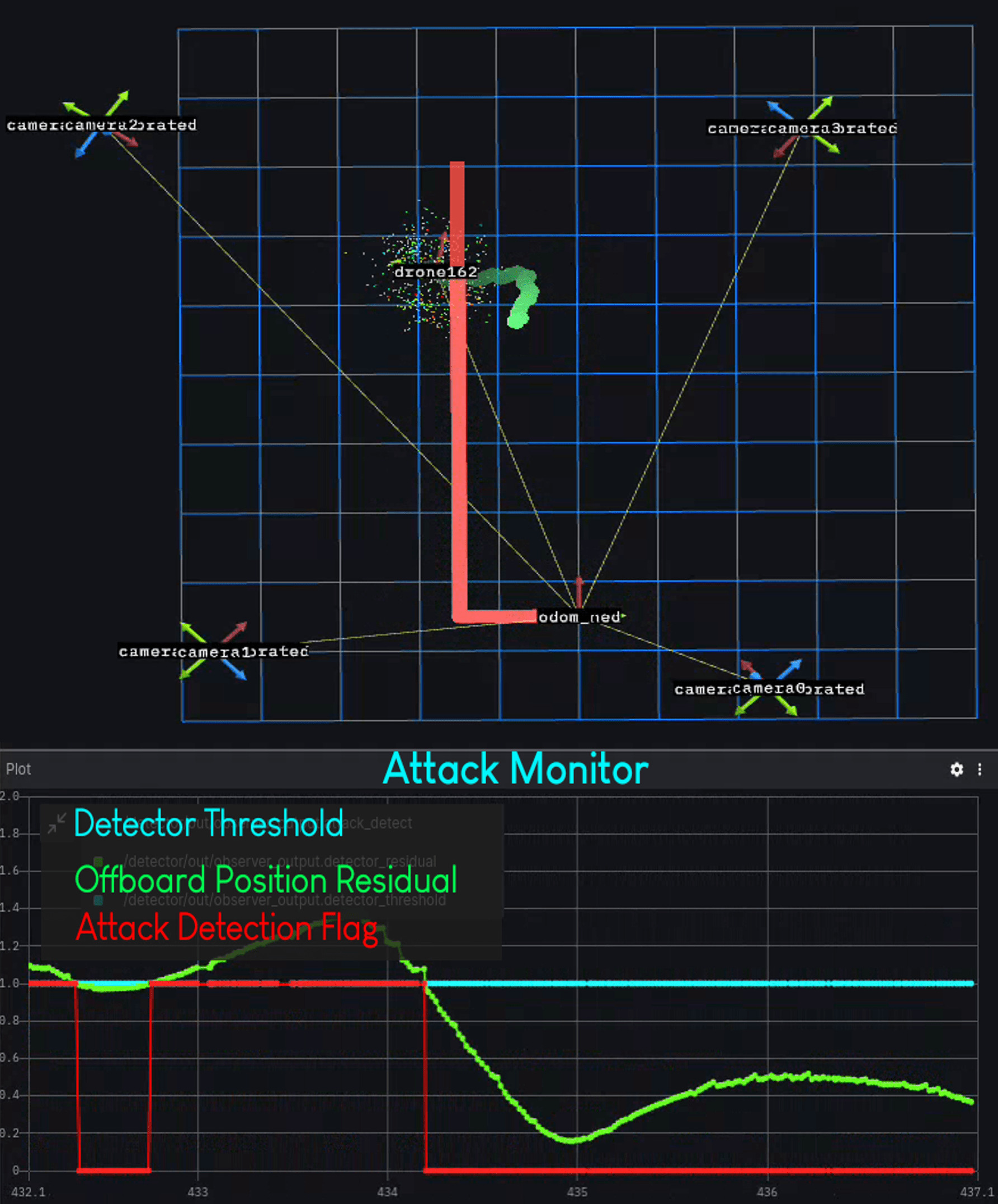}
      \caption{(Top Down) Particle filter-based UAV detection and tracking used as external measurements for attack detection. (Run-time Monitor) A linear observer-based detector that combines onboard and external measurements to detect and mitigate attacks.}
    \end{subfigure}
    \caption{Application of the proposed framework for test and validation of UAVs: Assessing the impacts of GNSS Meaconing attacks on the actual vehicle, and validating the use of external measurements for attack detection and mitigation.}
\end{figure*}

Under nominal operation, the test vehicle performs an autonomous waypoint navigation mission using PX4's offboard control mode. We then launch an emulated GNSS replay or meaconing attack on the vehicle, injecting false data into the GNSS measurements. The details of GNSS replay/meaconing attacks are presented in Sec. \ref{subsec:meacon}. The vehicle drifts off-course from its desired trajectory without triggering the onboard fault monitoring system, as shown in Fig. \ref{fig:experiment_a}. The PX4 Autopilot uses a chi-squared residual test for onboard fault monitoring \cite{px4ekf2020}. The residual signal represents the difference between the measurements and the predicted dynamics of the system. Under nominal conditions, the residual signal follows a normal distribution, and its energy (i.e., residual squared) follows a chi-squared distribution.
The chi-squared residual test compares the residual energy against a threshold to flag a faulty measurement signal. The attack is designed to deviate the UAV from the reference trajectory while maintaining the residuals below the detection threshold. 

Once the onboard sensing is compromised, a possible method for attack detection is to fuse external measurement redundancies (if available) along with the compromised onboard measurements. However, such an approach is only feasible if secure external measurements are available. Even when such external measurements are available, combining multi-rate and asynchronous measurements becomes a challenging problem. To address these issues, we leverage the idea of the lifting technique for residual generation and attack detection \footnote{As the main focus of this work is developing a mixed-reality framework for UAV cybersecurity, we direct interested readers to \cite{sun2022attack} for more details on lifting techniques for attack detection and mitigation}. We implement a linear observer-based fault detector for attack detection that fuses the external measurements with the onboard measurements to detect stealthy GNSS attacks. Section \ref{subsec:off_detect} highlights the details of the attack detector.

For demonstration, we use a camera network to provide secure external position measurements for detecting stealthy GNSS meaconing attacks. The details of the camera network are explained in Sec. \ref{subsec:visnet}. In real-world applications, such external measurements can be leveraged from existing infrastructure in urban environments; for example, camera surveillance systems for traffic monitoring or radar systems for air traffic control can be utilized to detect and track drones in urban environments. Finally, we demonstrate measurement reconfiguration where the vehicle fuses the external position measurements with the compromised GNSS measurement shown in Fig. \ref{fig:experiment_b} to correct its path. This is done by publishing the external position measurements to the visual odometry topic in PX4 once the attack detector detects the attack.
\subsection{GNSS Meaconing Attacks}
\label{subsec:meacon}
We demonstrate the false data injection attacks on GNSS measurements by designing GNSS replay/meaconing attacks. Such attacks have been demonstrated in \cite{motallebighomi2023location} \cite{lenhart2021relay} to be successful even against encrypted GNSS communication, i.e., Galileo's Open Service Navigation Message Authentication (OS-NMA). In our demonstration, we assume that an attacker launches a rogue spoofer drone equipped with a GNSS receiver and a radio transmitter to replay its received GNSS messages to the victim drone. The spoofer drone flies close to the victim drone, satisfying the transmitter power constraints and directionality conditions \cite{kerns2014unmanned}. The victim drone accepts the false data and starts following the trajectory guided by the spoofer drone. This attack is emulated using Gazebo by adding a spoofer drone inside the simulation environment. It follows a trajectory designed such that the onboard fault detection algorithm does not flag the rebroadcasted signals, i.e., residuals below a predefined threshold value. 

\subsection{Distributed Camera Network}
\label{subsec:visnet}
We utilized our previous work on a distributed camera network for UAV detection and tracking \cite{yang2022target}. It combines CNN-based object detection and particle filter tracking to detect and track multiple targets. The camera network provides external position measurements, which are subsequently used by the attack detector module to detect the presence of the emulated GNSS meaconing attack. It is implemented by creating an ad-hoc mesh network of four distributed camera nodes, each consisting of an UP Squared Edge Series with an Intel Atom x7-E3950 single-board computer and an Arducam 1080p USB serial camera.
\subsection{Offboard Attack Detector}
\label{subsec:off_detect}
We implement the linear observer-based fault detector presented in \cite{sun2022attack} to fuse the onboard and external position measurements. As mentioned earlier, the external measurements are obtained from the distributed camera network in our demonstration. The attack detector utilizes a lifting approach, i.e., augmentation of the states and measurements for residual generation. 
It has been shown to be effective for attack detection in the presence of multi-rate and asynchronous measurements.      

\section{Conclusion}
\label{sec:conclusion}
In this paper, we proposed a high-fidelity Mixed Reality sensor emulation framework for testing and evaluating UAVs against false data injection attacks. Our architecture allows an MRiTL experiment combining the physics of the actual vehicle with virtually emulated sensor measurements. We demonstrated our framework on an actual vehicle by generating an emulated GNSS meaconing attack. We also validated the use of external measurements to detect and mitigate the impact of such attacks using our framework. In the future, we will demonstrate the effectiveness of the proposed framework for testing and validating UAV swarms while adding real and virtual vehicles in tandem. This will allow us to model the large-scale interaction of swarms and assess the performance of fault detection, mitigation, and isolation algorithms through MRiTL experiments.  

\bibliographystyle{ieeetr}
\bibliography{IEEEabrv,bibs} 

\end{document}